\documentclass[conference]{IEEEtran}
\ifCLASSINFOpdf
\else
\fi
\usepackage{amsmath}
\usepackage{graphicx}

\usepackage{booktabs}
\usepackage{multirow}
\usepackage{graphicx}

\begin{document}
%
\title{Divide and Restore: A Modular Task-Decoupled Framework for Universal Image Restoration}



%
\author{\IEEEauthorblockN{J. Wiekiera\IEEEauthorrefmark{1} and
M. Żur\IEEEauthorrefmark{1}}
\IEEEauthorblockA{\IEEEauthorrefmark{1}Faculty of Automatic Control, Electronics and Computer Science\\
Silesian University of Technology, Gliwice, Poland\\
Email: jw302114@student.polsl.pl, mz302271@student.polsl.pl}}


\maketitle

\begin{abstract}
Restoring images affected by various types of degradation, such as noise, blur, or improper exposure, remains a~ significant challenge in computer vision. While recent trends favor complex monolithic all-in-one architectures, these models often suffer from negative task interference and require extensive joint training cycles on high-end computing clusters. In this paper, we propose a modular, task-decoupled image restoration framework based on an explicit diagnostic routing mechanism. The architecture consists of a lightweight Convolutional Neural Network (CNN) classifier that evaluates the input image and dynamically directs it to a specialized restoration node. A key advantage of this framework is its model-agnostic extensibility: while we demonstrate it using three independent U-Net experts, the system allows for the integration of any restoration method tailored to specific tasks. By isolating reconstruction paths, the~framework prevents feature conflicts and significantly reduces training overhead. Unlike monolithic models, adding new degradation types in our framework only requires training a~single expert and updating the router, rather than a full system retraining. Experimental results demonstrate that this computationally accessible approach offers a scalable and efficient solution for multi-degradation restoration on standard local hardware. The code will be published upon paper acceptance.
\end{abstract}


\IEEEpeerreviewmaketitle

%

\section{Introduction}
Digital images play an important role in many areas such as photography, social media, medical imaging, and surveillance systems. However, images are often affected by various types of degradation caused by poor lighting conditions, camera limitations, environmental factors, or data transmission errors. Common image quality problems include blur, noise, and overexposure, which can significantly reduce visual quality and limit the usability of images.

Image restoration aims to recover clean and visually pleasing images from corrupted inputs. Traditional image processing methods are based on handcrafted filters and mathematical models, which often struggle to generalize between different types of degradation and real-world conditions. In recent years, deep learning approaches have shown strong performance in~image restoration tasks by learning complex mappings directly from data \cite{zhang2017beyond}.

To address multiple types of degradation within a single system, recent trends have shifted towards unified, ``all-in-one'' architectures, documented in recent comprehensive surveys \cite{jiang2024survey}. However, these monolithic models are often built as rigid entities that lack modularity, forcing highly heterogeneous tasks to share the same feature extraction backbone. This often leads to negative task interference and substantial computational overhead. A major drawback of such designs is~the~necessity to retrain the entire network from scratch on massive datasets whenever a new type of degradation needs to~be~supported. This global retraining is not only computationally expensive but also poses a high risk of catastrophic forgetting, where the model loses its previously acquired restoration capabilities while adapting to new data. Consequently, maintaining and extending these large-scale systems requires immense processing power and long development~cycles.

In contrast, Convolutional Neural Networks (CNNs), especially encoder–decoder architectures, have become widely used for image-to-image translation problems \cite{isola2017image} due to their linear computational complexity. Among them, the U-Net architecture \cite{ronneberger2015u} has gained popularity due to its ability to preserve spatial information through skip connections. This makes U-Net particularly suitable and highly efficient for restoring fine image details that are often lost during degradation.

In this work, we propose a task-decoupled image restoration framework based on explicit routing to leverage this computational efficiency. The proposed architecture introduces a lightweight classification network that evaluates the input image and dynamically assigns it to one of three specialized deep learning experts designed to handle blur, overexposure, and noise. Each U-Net model is trained separately to restore images affected by a specific type of corruption. 

This modular approach enables targeted image recovery and completely prevents the negative transfer associated with shared backbones. Most importantly, by dividing the problem into isolated tasks, the framework significantly accelerates the~training process. Each expert can be trained independently on standard local hardware in a fraction of the time required for complex monolithic models, despite maintaining a comparable active parameter count during inference. This study demonstrates that a decoupled, routing-based U-Net architecture is an effective, computationally efficient, and accessible solution for restoring corrupted images in practical multi-degradation applications.

\section{Related Work}

\subsection{Deep Learning in Image Restoration}
Deep learning has fundamentally shifted the paradigm of~image restoration from handcrafted filters to data-driven representations. Comprehensive reviews by Zhai et al. \cite{zhai2023comprehensive} confirm that data-driven models generalize significantly better to~real-world degradations than traditional mathematical models (e.g., Wiener \cite{wiener1949extrapolation} filtering), which are often limited by~ rigid~assumptions.

Building on this, a pivotal advancement was introduced by~Zhang et al. with the DnCNN architecture \cite{zhang2017beyond}. The authors demonstrated that training a network to predict the \textit{residual noise map}, rather than the clean image directly, significantly accelerates convergence and improves denoising performance compared to classical benchmarks like BM3D \cite{dabov2007image}. 

\subsection{Blind vs. Non-Blind Restoration}
A critical challenge in restoration is handling variable degradation types without explicit guidance (``blind'' restoration). In the SRMD framework, Zhang et al. \cite{zhang2018learning} demonstrated that blind models suffer from severe performance drops when input statistics deviate from the training distribution (domain shift). They proved that providing the network with explicit degradation parameters (a \textit{non-blind} approach) significantly boosts restoration quality. 

While SRMD achieves this via dimensionality stretching (concatenating degradation maps to the input), our system implements the non-blind philosophy through a system-level \textit{routing mechanism}. By first classifying the degradation and directing the image to a specialized expert, we ensure that the~restoration model operates in a non-blind regime, effectively bypassing the need for the network to implicitly estimate the degradation type.

\subsection{Degradation Classification and Experts}
The concept of explicitly modeling degradation types originates from Wang et al. \cite{wang2020deep}, who introduced the \textit{Deep Degradation Prior}. They demonstrated that different degradations (noise, blur) form distinct clusters in the feature space. While Wang et al. used this primarily to improve \textit{semantic classification} (object recognition), we repurpose this finding for restoration, utilizing the classifier to actively \textit{route} the input to a specialized repair model.

The validity of such a modular strategy is corroborated by recent state-of-the-art advancements, notably the unified DeMoE architecture introduced by Feijoo et al. (2025) \cite{feijoo2025towards}. These works confirm that decomposing complex restoration tasks via \textit{Mixture-of-Experts (MoE)} yields better generalization than monolithic ``All-in-One'' networks. 

However, our system introduces a critical advantage in terms of extensibility. Unlike monolithic MoE architectures where adding a new degradation type often necessitates retraining the~shared parameters (risking catastrophic forgetting), our system employs a modular \textit{Hard Gating} strategy. New experts can be trained independently and integrated simply by updating the routing logic, without altering the weights of existing experts.

\subsection{Architectural Backbone}
For the restoration experts, we adopt the U-Net architecture. This choice is grounded in the findings of Isola et al. (Pix2Pix) \cite{isola2017image}, who investigated architectures for general \textit{Image-to-Image Translation}. They demonstrated the critical role of \textit{skip connections} in bypassing the information bottleneck of standard Encoder-Decoder networks. These connections allow low-level features (such as edges and textures) to propagate directly to~the~output, which is essential for preserving structural fidelity in restoration tasks.

\subsection{Overexposure Correction}
While denoising and deblurring are extensively studied, \textit{overexposure correction} remains an under-explored niche in~unified frameworks compared to low-light enhancement \cite{zhai2023comprehensive}. Recovering overexposed areas is an \textit{ill-posed problem} involving the reconstruction of information physically lost due to sensor saturation (clipping). 

While some standalone methods exist \cite{afifi2021learning}, most unified benchmarks focus on recovering signal from noise or darkness. Our system addresses this gap by incorporating a dedicated expert trained to hallucinate missing textures in clipped regions, treating overexposure as a distinct degradation class rather than a simple intensity adjustment.

\subsection{MoE vs. Modular Framework}
Recent advancements in all-in-one restoration, such~as~MEASNet \cite{yu2025measnet} and DaAIR \cite{zamfir2024efficient}, attempt to~unify restoration tasks by embedding a Mixture-of-Experts (MoE) mechanism directly into the network’s hidden layers. In these designs, experts function as internal sub-networks (e.g., MLP blocks or attention-based modules) situated within a shared monolithic backbone. 

A significant drawback of this approach is the requirement for extensive \textit{joint training}. As documented in \cite{yu2025measnet}, these models often require between 120 to 150 epochs of joint optimization on combined datasets to synchronize the interdependent weights of the shared encoder and decoder. This architecture prevents true modularity; any attempt to extend the system necessitates a full re-training of the entire monolith to avoid catastrophic forgetting.

\section{Methodology}

  \subsection{System Overview}                                                                                                                                                                                                      
\begin{figure*}[t]                                                                                                                                                                                                                
    \centering                                                                                                                                                                                                                      
    \includegraphics[width=\textwidth]{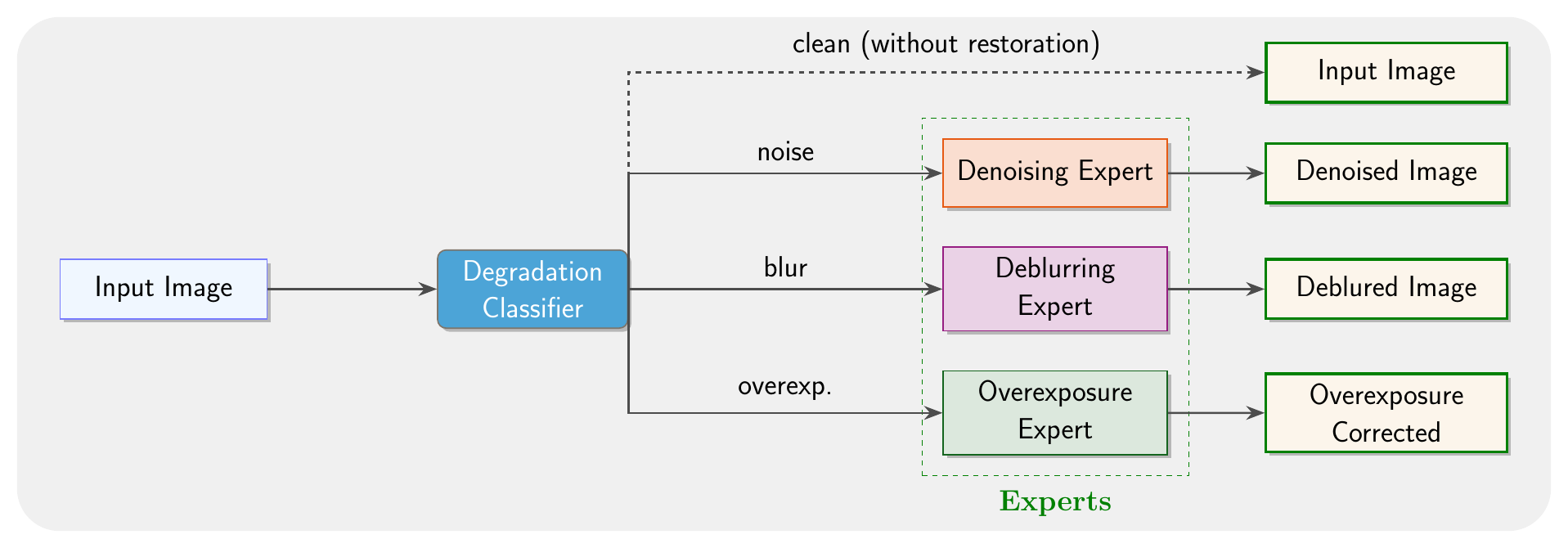}                                                                                                                                                                            
    \caption{Overview of the DaR-Net framework. The input image
    is first evaluated by a lightweight CNN classifier, which routes it to
    the corresponding U-Net expert for task-specific restoration.}
    \label{fig:pipeline}
  \end{figure*}

  We propose \textit{Divide-and-Restore} (DaR-Net), a modular image restoration
  pipeline based on a Mixture-of-Experts (MoE) architecture \cite{shazeer2017outrageously}. As illustrated                      
  in Fig.~\ref{fig:pipeline}, the system consists of two decoupled components:
  a degradation classifier that acts as a router, and a set of independent
  restoration experts. Given a~corrupted image, the classifier first identifies
  the~degradation type, then routes the image to the corresponding expert for
  restoration. This design contrasts with all-in-one restorers, where a single
  network must simultaneously handle all degradation types.

\subsection{Degradation Classifier}
The routing mechanism is implemented using a~lightweight CNN acting as a discrete gate, mapping the input image $I$ to a single categorical label $y \in \{\text{Clean}, \text{Noise}, \text{Blur}, \text{Overexposure}\}$ via argmax over the output logits. Unlike adaptive selection schemes that compute weighted combinations of experts \cite{puigcerver2024soft}, this hard-gating strategy activates only one specialized expert per inference pass, by design avoiding redundant computations and the blending of incompatible feature representations.

\subsection{Restoration Experts}
Each restoration expert $E_i$ is built upon a U-Net architecture with skip connections between encoder and decoder. The~output is passed through a sigmoid activation, yielding pixel values directly in the $[0,1]$ range. All three experts share the same architecture and are trained independently on task-specific data with a hybrid loss combining L1 and SSIM \cite{wang2004image}:
\begin{equation}
  \mathcal{L} = \alpha\,(1 - \text{SSIM}(\hat{I}, I)) + (1-\alpha)\,\|\hat{I} - I\|_1,
\end{equation}
where $\hat{I}$ is the restored image and $I$ is the clean reference. This formulation balances pixel-level fidelity (L1) with perceptual quality (SSIM). 

Importantly, the framework is agnostic to the expert implementation. The U-Net instantiation presented here can be~transparently replaced by classical signal-processing methods or other deep architectures without modifying the routing~mechanism.

\subsection{Extensibility}
The modular architecture is inherently extensible: adding support for a new degradation type requires training one additional expert and running one additional continual learning phase on the classifier, leveraging experience replay \cite{rebuffi2017icarl} to~preserve performance on previously learned degradation types. Crucially, existing expert models require no retraining, as the router is updated incrementally rather than from scratch. This stands in contrast to all-in-one architectures, which require full retraining when the degradation set changes.

\section{Experimental Evaluation}  

\subsection{Experimental Setup}

The degradation classifier is trained on STL-10 \cite{coates2011analysis} following a three-phase continual learning curriculum, with each phase introducing one additional degradation type: Phase 1 covers clean and noisy images, Phase 2 adds blur, and Phase 3 adds overexposure. A 70/10/20 train/validation/test split is~applied, with batch size 256 and an initial learning rate of~$10^{-3}$, reduced by a factor of 0.5 on validation loss plateau, and early stopping after 7 non-improving epochs (maximum 100 epochs). The three phases converged in 21, 28, and 41 epochs respectively, with a total training time of approximately 14 minutes.

The U-Net restoration experts are trained on the DIV2K \cite{agustsson2017ntire} training set (800 high-resolution images). For each epoch, 50 random $128\times128$ crops are extracted per image, yielding approximately 40\,000 training samples, with a 90/10 train/validation split. Experts are optimized using Adam \cite{kingma2014adam} with the same learning rate schedule and batch size 128. Degradations are applied synthetically on-the-fly within the~following magnitude ranges: Gaussian noise $\sigma \in [15, 50]$, Gaussian blur $\sigma \in [0.5, 3.0]$, and overexposure $\gamma \in [1.3, 2.0]$. Individual experts required between 12 and 36 training epochs (noise: 35, blur: 36, overexposure: 12), with training times of~ approximately 9--11 hours per model. All experiments were conducted on a single NVIDIA RTX 4070 GPU.

\subsection{Comparison with State-of-the-Art}                                                                                 
Table~\ref{tab:denoising_bsd68_final} reports Gaussian denoising results on the BSD68 benchmark at three standard noise levels ($\sigma \in \{15, 25, 50\}$). \textit{DaR-Net} achieves competitive performance relative to state-of-the-art all-in-one restorers, despite being a modular system trained without any joint multi-task optimization. Notably, our method achieves competitive average SSIM, and the highest SSIM at the hardest noise level ($\sigma=50$). The~performance gap in PSNR compared to the best-performing baseline (MEASNet) is within 0.6 dB. We attribute this primarily to~the~ difference in training regime: competing methods undergo expensive joint multi-task training on large combined datasets, while our experts are trained independently on single-task data. Additionally, transformer-based backbones employed by MEASNet and MoCE-IR benefit from global self-attention, which captures long-range dependencies unavailable to convolutional U-Nets. The primary contribution of our framework is not peak restoration performance, but rather modularity, task isolation, and extensibility, properties evaluated in Section~\ref{sec:continual}.

\begin{table*}[t]
\centering
\renewcommand{\arraystretch}{1.3}
\setlength{\tabcolsep}{10pt}
\caption{Quantitative evaluation on the BSD68 \cite{martin2001database} dataset for Gaussian denoising.
\textbf{Bold} values indicate the best performance per column.}
\label{tab:denoising_bsd68_final}
\resizebox{\textwidth}{!}{%
\begin{tabular}{lcccccccc}
\toprule
\multirow{2}{*}{\textbf{Method}}
& \multicolumn{2}{c}{\textbf{$\sigma=15$}}
& \multicolumn{2}{c}{\textbf{$\sigma=25$}}
& \multicolumn{2}{c}{\textbf{$\sigma=50$}}
& \multicolumn{2}{c}{\textbf{Average}} \\
\cmidrule(lr){2-3}\cmidrule(lr){4-5}\cmidrule(lr){6-7}\cmidrule(l){8-9}
& \textbf{PSNR} & \textbf{SSIM}
& \textbf{PSNR} & \textbf{SSIM}
& \textbf{PSNR} & \textbf{SSIM}
& \textbf{PSNR} & \textbf{SSIM} \\ \midrule
AirNet \cite{li2022all}
& 33.920 & 0.933 & 31.260 & 0.888 & 28.000 & 0.797 & 31.060 & 0.873 \\
Restormer \cite{zamir2022restormer}
& 31.960 & ---   & 29.510 & ---   & 26.620 & ---   & 29.363 & ---   \\
PromptIR \cite{potlapalli2023promptir}
& 33.980 & 0.933 & 31.310 & 0.888 & 28.060 & 0.799 & 31.117 & 0.873 \\
DaAIR \cite{zamfir2024efficient}
& 33.920 & 0.930 & 31.260 & 0.884 & 28.000 & 0.792 & 31.060 & 0.869 \\
MoCE-IR \cite{zamfir2025complexity}
& 34.080 & 0.933 & 31.420 & 0.888 & 28.160 & 0.798 & 31.220 & 0.873 \\
MEASNet \cite{yu2025measnet}
      & \textbf{34.120} & \textbf{0.935} & \textbf{31.460} & \textbf{0.892} & \textbf{28.190} & 0.803 & \textbf{31.257} & \textbf{0.877} \\ \midrule
    \textbf{DaR-NET (ours)}
      & 33.550 & 0.933 & 31.030 & 0.890 & 27.650 & \textbf{0.804} & 30.743 & 0.875 \\
\bottomrule
\end{tabular}%
}
\end{table*}

\subsection{Continual Learning Evaluation}                   \label{sec:continual}                                        
Table~\ref{tab:continual} reports restoration quality across the three continual learning phases for a representative degradation level per task. In Phase~1, the classifier is trained on clean and noisy images only, while blur and overexposure inputs are misrouted, yielding corrupted-level metrics. Phase~2 introduces the blur expert, immediately recovering deblurring performance, while noise results remain unchanged, confirming that experience replay \cite{rebuffi2017icarl} effectively prevents catastrophic forgetting. Phase~3 adds the overexposure expert, completing the full system, with no degradation in previously learned tasks.

\begin{figure}[t]
  \centering
  \includegraphics[width=\columnwidth]{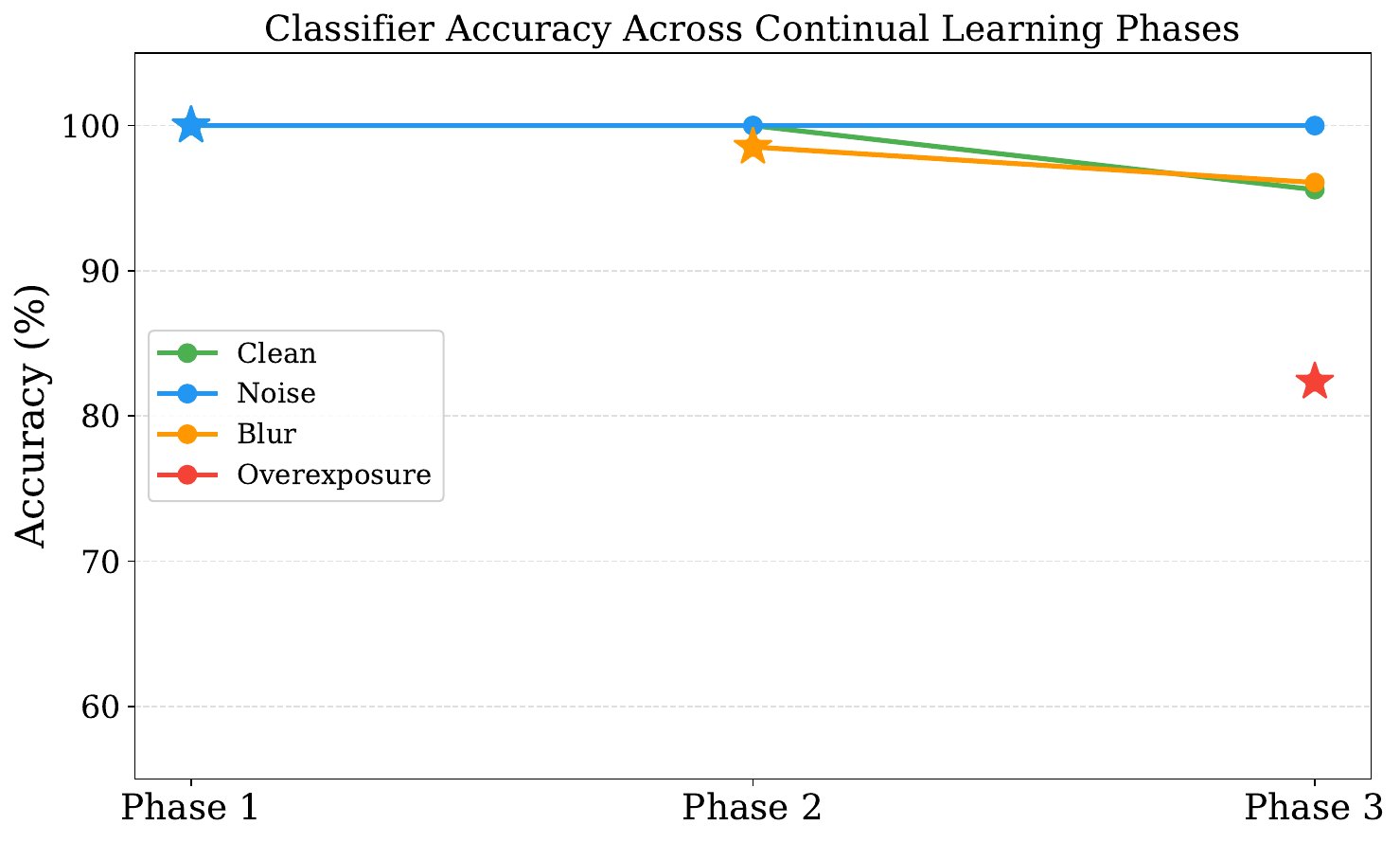}                                                                                                                                                             
  \caption{Classifier accuracy per degradation class across continual learning phases, evaluated on BSD68. Stars mark the phase in which each class was introduced. Clean has no star as it corresponds to unmodified images present throughout all phases. Experience replay prevents accuracy drops on previously learned classes.}
  \label{fig:continual_lines}
\end{figure}

Figure~\ref{fig:continual_lines} illustrates classifier validation accuracy across all three phases. The system achieves 100\%, 99.38\%, and 95.77\% accuracy after Phases~1, 2, and 3 respectively, demonstrating that incremental classifier updates preserve discriminative performance on earlier degradation types. Results for Gaussian blur and overexposure correction are not directly comparable to the all-in-one baselines in Table~\ref{tab:denoising_bsd68_final}, as these methods were not designed or trained to handle these specific degradation types. We therefore report our results independently as a reference for future work on unified multi-degradation benchmarks.

\begin{table}[t]
\centering
\renewcommand{\arraystretch}{1.3}
\footnotesize 
\setlength{\tabcolsep}{2.5pt} 

\caption{Restoration quality (PSNR~[dB] / SSIM) across continual learning phases
on BSD68 \cite{martin2001database}. Each phase introduces one new degradation type. Corrupted denotes
the unrestored input. \textbf{Bold} marks the phase that first activates each expert.}
\label{tab:continual}

\begin{tabular}{llcccc}
\toprule
\textbf{Task} & \textbf{Level} & \textbf{Corrupted} & \textbf{Phase 1} & \textbf{Phase 2} & \textbf{Phase 3} \\
\midrule
Noise    & $\sigma=25$   & 20.48 / .391 & \textbf{31.03 / .890} & 31.03 / .890 & 31.03 / .890 \\
Blur     & $\sigma=1.5$  & 26.07 / .744 & 26.07 / .744 & \textbf{30.20 / .906} & 30.20 / .906 \\
Overexp. & $\gamma=1.7$  & 11.98 / .768 & 11.98 / .768 & 11.98 / .768 & \textbf{20.44 / .891} \\
\bottomrule
\end{tabular}
\end{table}

 \subsection{Qualitative Results}                          

\begin{figure}[t]                                                                                                                                                                                                                 
  \centering                                                                                                                                                                                                                        
  \includegraphics[width=\columnwidth]{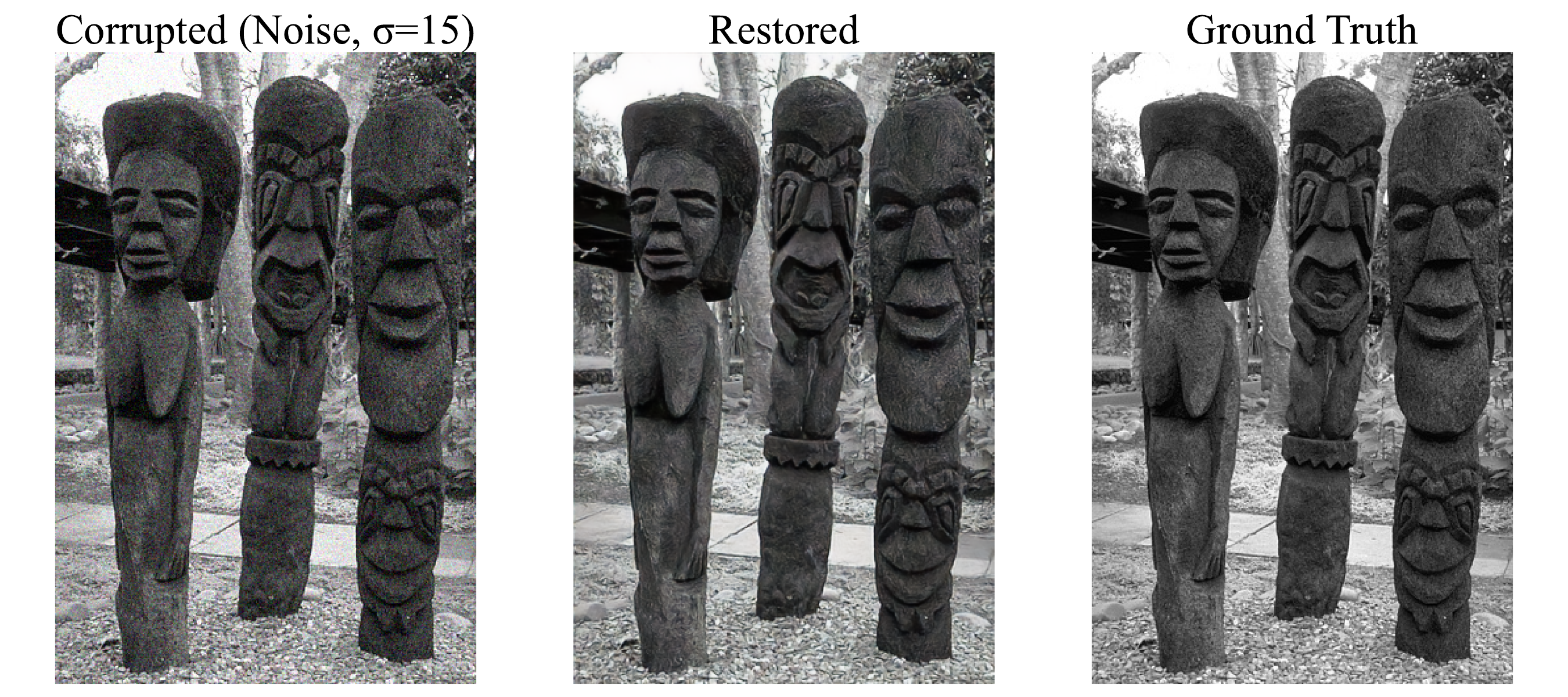}\\[0pt]       
  \includegraphics[width=\columnwidth]{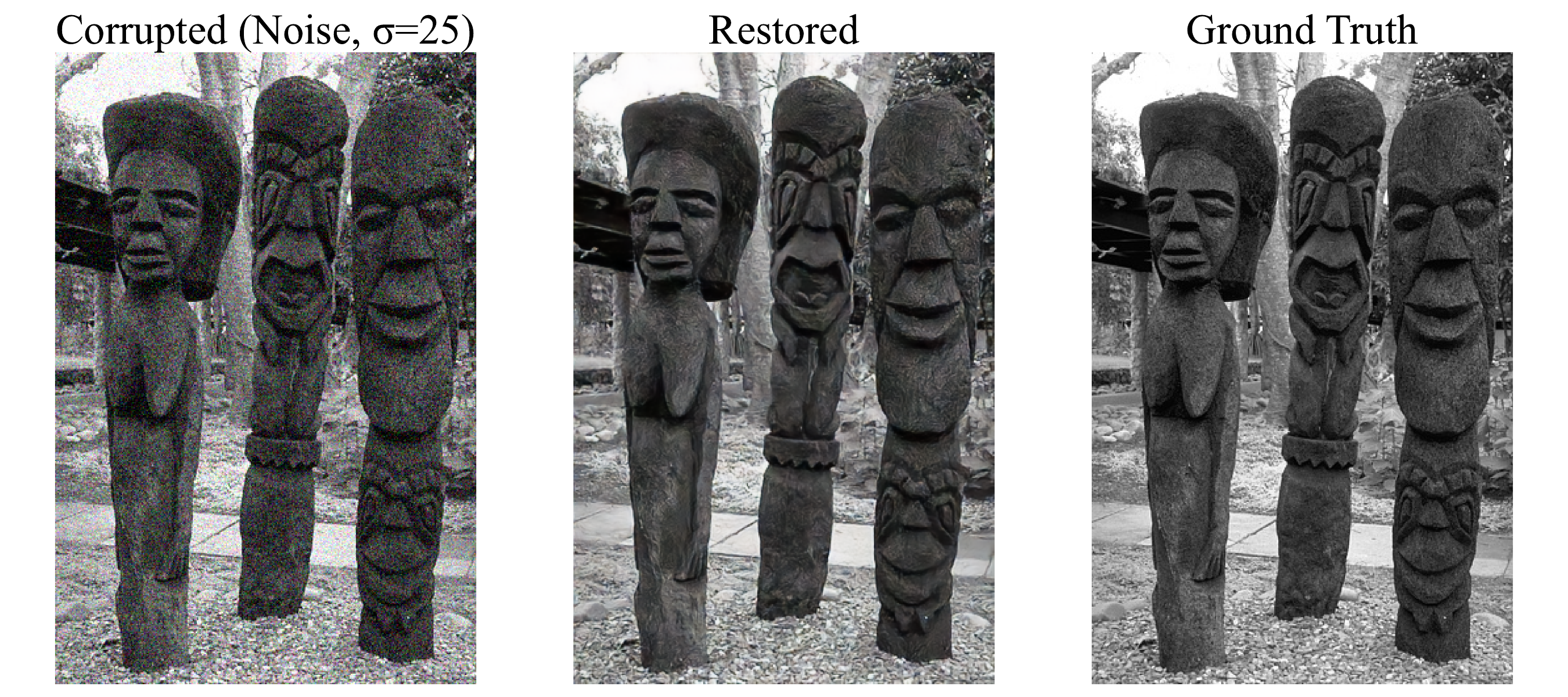}\\[0pt]
  \includegraphics[width=\columnwidth]{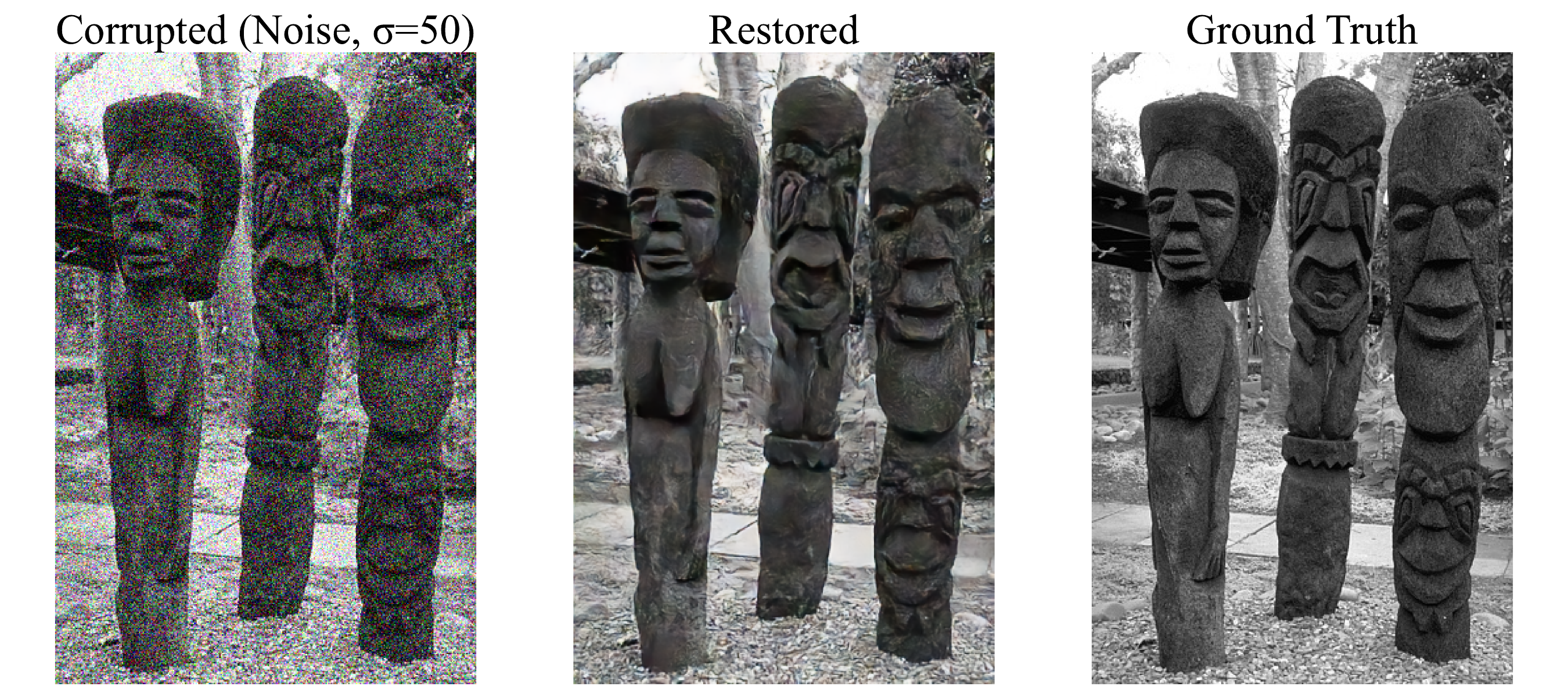}
  \caption{Gaussian noise denoising on a BSD68 test image across three noise levels.
  Each row shows the corrupted input, DaR-Net restoration, and ground truth,
  with metrics reported as PSNR~[dB]\,/\,SSIM\,/\,LPIPS~\cite{zhang2018unreasonable}$\downarrow$.                                                                                                                                   
  \textbf{Top} ($\sigma{=}15$): 30.61\,/\,0.908\,/\,0.153.
  \textbf{Middle} ($\sigma{=}25$): 27.95\,/\,0.841\,/\,0.218.                                                                                                                                                                       
  \textbf{Bottom} ($\sigma{=}50$): 24.60\,/\,0.702\,/\,0.359.}
  \label{fig:qualitative_noise}
  \end{figure}

\begin{figure}[t]
  \centering
  \includegraphics[width=\columnwidth]{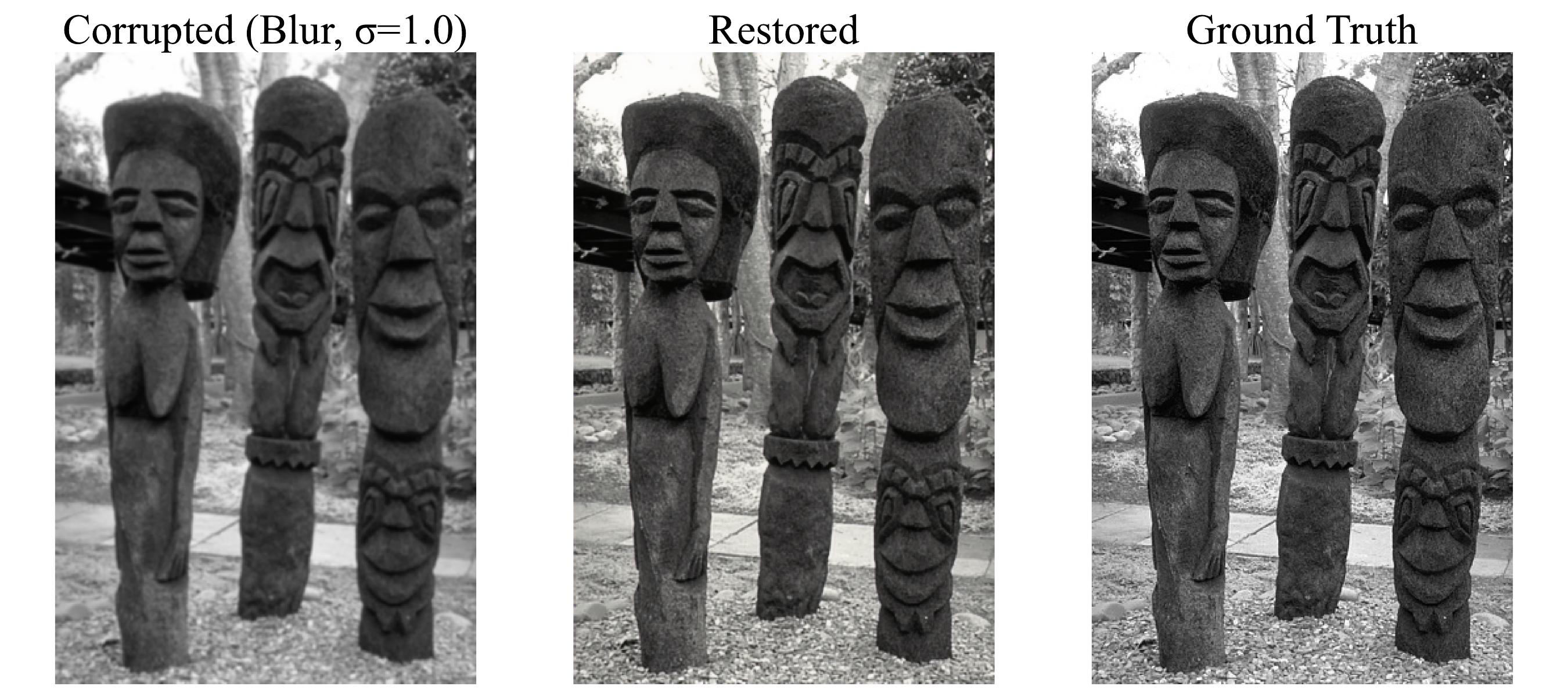}\\[0pt]
  \includegraphics[width=\columnwidth]{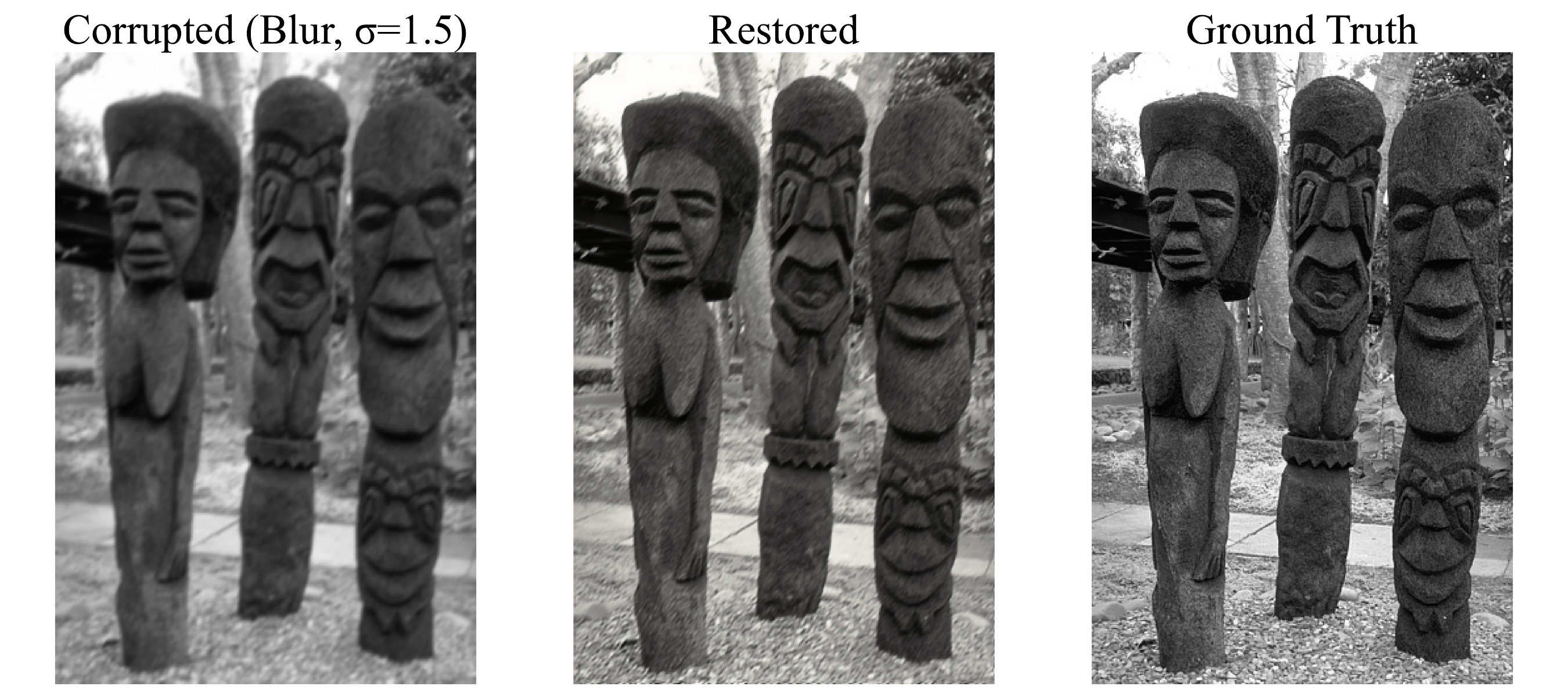}\\[0pt]
  \includegraphics[width=\columnwidth]{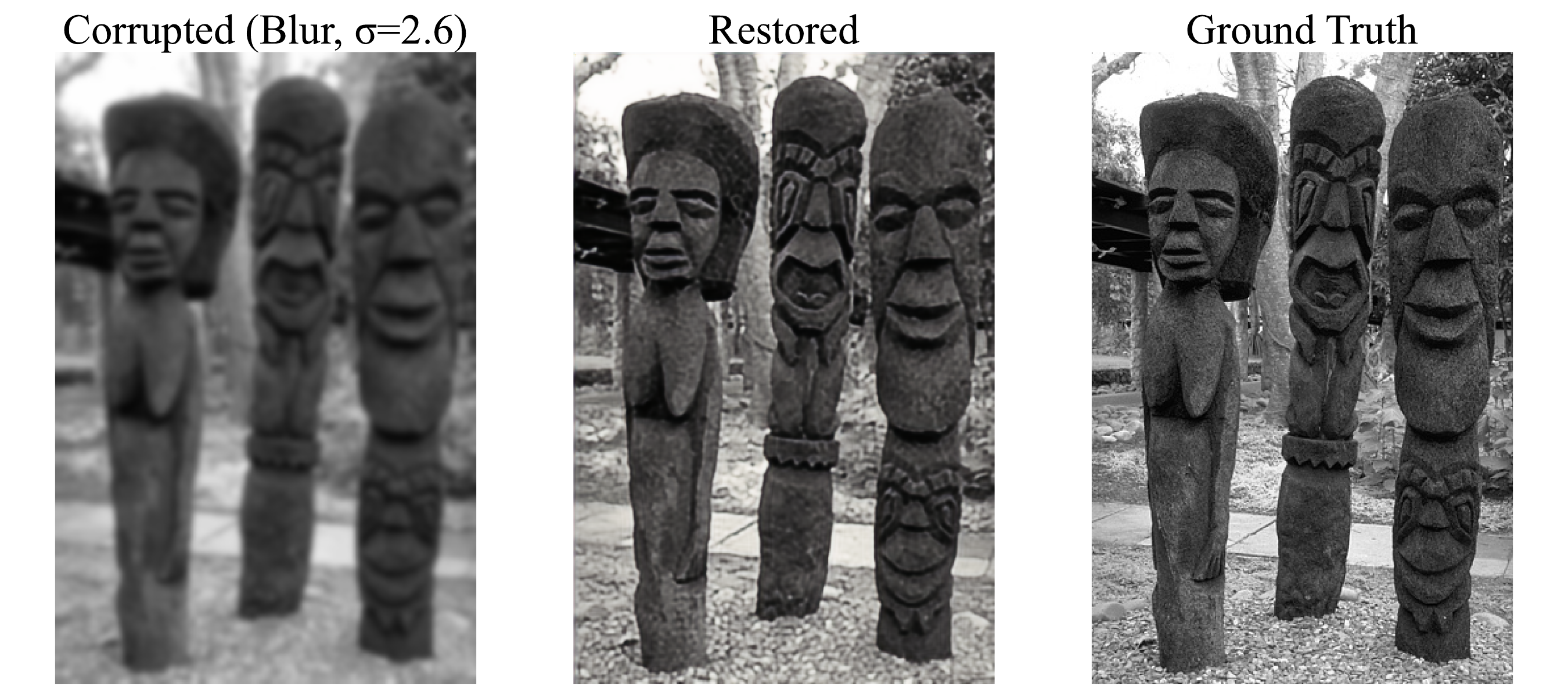}
    \caption{Gaussian blur restoration on a BSD68 test image across three blur levels.
  Each row shows the corrupted input, DaR-Net restoration, and ground truth,
  with metrics reported as PSNR~[dB]\,/\,SSIM\,/\,LPIPS~$\downarrow$.
  \textbf{Top} ($\sigma{=}1.0$): 30.58\,/\,0.931\,/\,0.078.
  \textbf{Middle} ($\sigma{=}1.5$): 25.18\,/\,0.766\,/\,0.382.
  \textbf{Bottom} ($\sigma{=}2.6$): 24.05\,/\,0.702\,/\,0.344.}
  \label{fig:qualitative_blur}
  \end{figure}

\begin{figure}[t]
  \centering
  \includegraphics[width=\columnwidth]{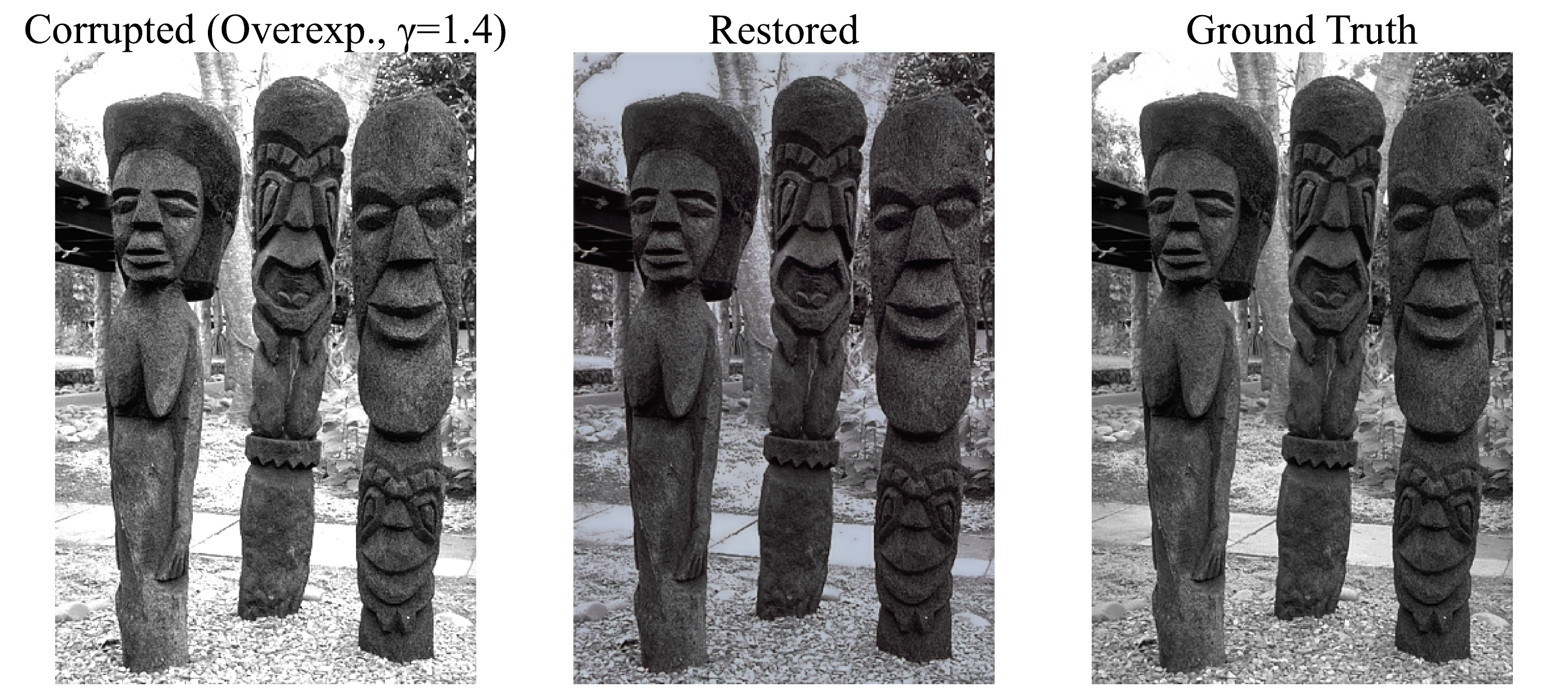}\\[0pt]
  \includegraphics[width=\columnwidth]{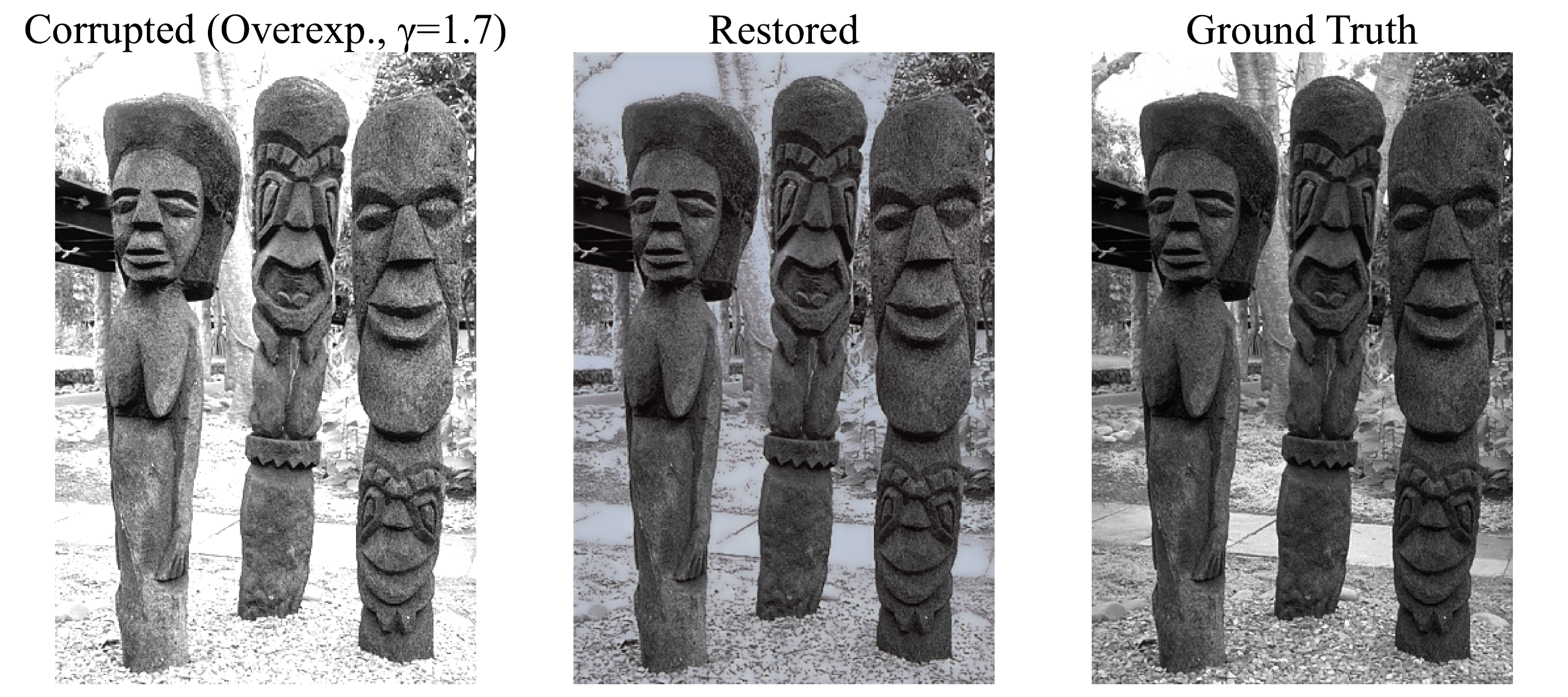}\\[0pt]
  \includegraphics[width=\columnwidth]{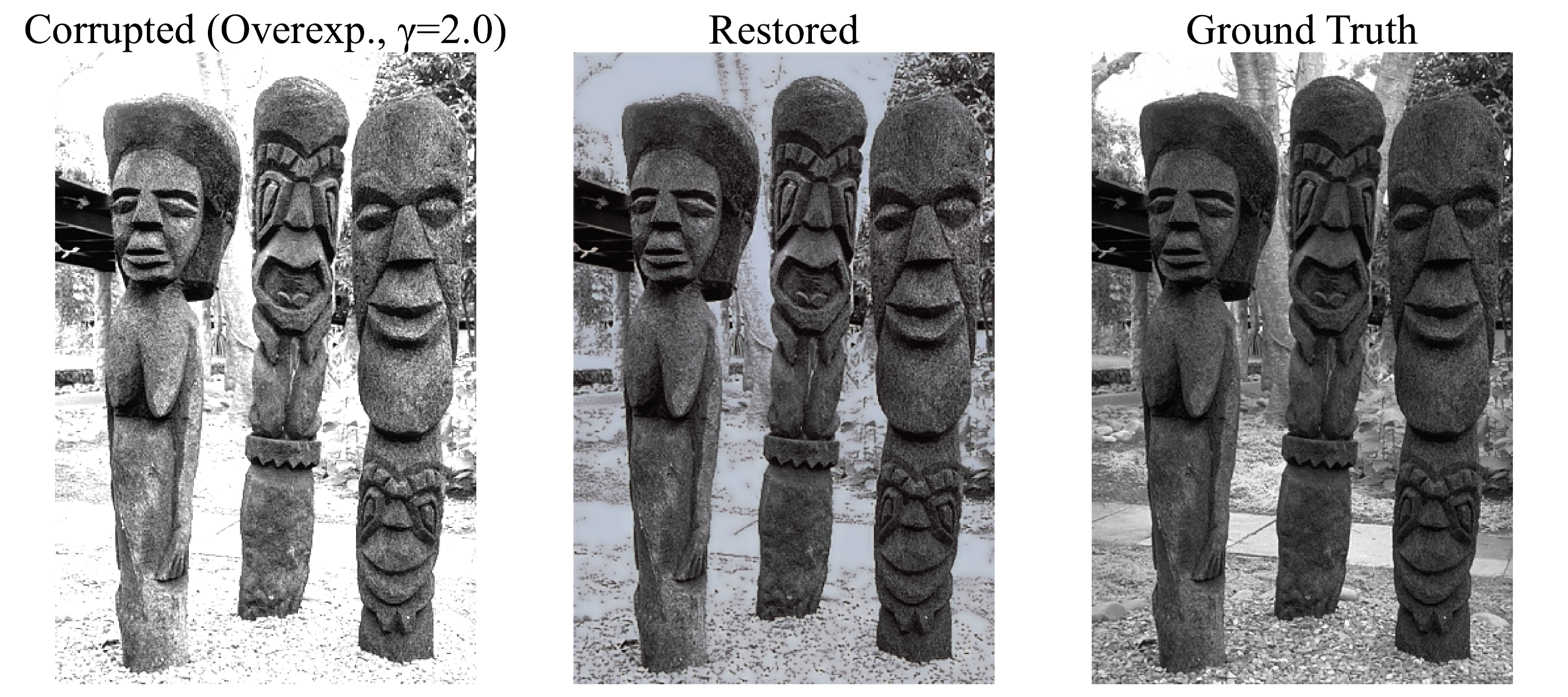}
    \caption{Overexposure correction on a BSD68 test image across three exposure levels.
  Each row shows the corrupted input, DaR-Net restoration, and ground truth,
  with metrics reported as PSNR~[dB]\,/\,SSIM\,/\,LPIPS~$\downarrow$.
  \textbf{Top} ($\gamma{=}1.4$): 24.22\,/\,0.963\,/\,0.068.
  \textbf{Middle} ($\gamma{=}1.7$): 22.58\,/\,0.908\,/\,0.126.
  \textbf{Bottom} ($\gamma{=}2.0$): 20.50\,/\,0.850\,/\,0.181.}
  \label{fig:qualitative_overexposure}
  \end{figure}

Figures~\ref{fig:qualitative_noise}–\ref{fig:qualitative_overexposure} show representative restorations produced by~the full DaR-Net pipeline on a single BSD68 image across three severity levels for each degradation type.         

For Gaussian noise (Figure~\ref{fig:qualitative_noise}), the network suppresses grain while preserving fine texture across all tested levels. At $\sigma{=}15$ the network almost fully suppresses noise artefacts (PSNR~30.61~dB / SSIM~0.908); performance degrades gracefully with increasing noise, reaching PSNR~27.95~dB / SSIM~0.841 at $\sigma{=}25$ and PSNR~24.60~dB / SSIM~0.702~at~$\sigma{=}50$.  

Gaussian blur (Figure~\ref{fig:qualitative_blur}) is the most challenging task: at~ $\sigma{=}1.0$ the expert recovers fine detail well (PSNR~30.58~dB / SSIM~0.931), but some high-frequency sharpness is lost at~stronger blur levels (PSNR~25.18~dB / SSIM~0.766 at~$\sigma{=}1.5$; PSNR~24.05~dB / SSIM~0.702 at $\sigma{=}2.6$).

Overexposure correction (Figure~\ref{fig:qualitative_overexposure}) yields the largest absolute PSNR gains. Even at mild overexposure ($\gamma{=}1.4$, PSNR~24.22~dB / SSIM~0.963) the network restores natural contrast and colour balance; at severe clipping ($\gamma{=}2.0$) it still achieves PSNR~20.50~dB / SSIM~0.850 despite a +9.18~dB gain over the corrupted input.


\section{Conclusions}                                                                                                                                                                                                             
  We presented \textit{DaR-Net}, a modular image restoration
  framework based on hard-gated routing and independent task-specific experts.                                                                                                                                                      
  By decoupling degradation classification from restoration, the system avoids
  negative task interference inherent in monolithic all-in-one architectures                                                                                                                                                        
  and significantly reduces training overhead: each U-Net expert is trained
  independently on a single degradation type in approximately 9--11 hours on
  commodity hardware, compared to multi-day joint training cycles required by
  transformer-based unified restorers.

  Quantitative evaluation on BSD68 demonstrates competitive denoising
  performance relative to state-of-the-art all-in-one methods, with a
  performance gap of under 0.6~dB PSNR compared to the best baseline
  (MEASNet), despite no joint multi-task training. Continual learning
  experiments confirm that the classifier can be extended incrementally
  across three degradation phases with experience replay, preserving
  accuracy on~previously learned classes (100\%, 99.4\%, and 95.8\%
  after Phases 1--3 respectively) without retraining existing experts.

  The primary limitation of the current system is its assumption of a
  single dominant degradation per image; real-world images frequently
  exhibit mixed or spatially varying corruption. Future work should
  therefore focus on handling composite degradations, exploring
  confidence-based soft routing for ambiguous inputs, and investigating
  alternative backbone architectures for the restoration experts.
  The modular design of \textit{Divide-and-Restore} makes all of these
  extensions straightforward to integrate without modifying existing
  components.

\bibliographystyle{IEEEtran}
\bibliography{references}

\end{document}